\RequirePackage{fix-cm}
\documentclass[smallextended]{svjour3}       % onecolumn (second format)
\smartqed  % flush right qed marks, e.g. at end of proof
\usepackage{booktabs} % For formal tables
\usepackage{graphicx}
\usepackage{amsfonts}
\usepackage{dsfont}
\usepackage{graphicx, epsfig,amsmath,amssymb,latexsym,graphics,float}
\usepackage{setspace,subfig}
\usepackage{citesort}
\usepackage{float}
\usepackage{booktabs}
\usepackage{lipsum}
\usepackage{multicol}
\usepackage{url}
\usepackage{multirow}
\usepackage{algorithm}
\usepackage{algorithmic}
\begin{document}

\title{Particle Filtering Methods for Stochastic Optimization with Application to Large-Scale Empirical Risk Minimization
%\thanks{This work was partly supported by National key research and development plan of China (No.YFB2101704),
%the National Natural Science Foundation (NSF) of China (Nos. 61571238 and 61906099),
%and Scientific Research Foundation of Nanjing University of Posts and Telecommunications (No.NY218072).}
}

\author{Bin Liu
}

\institute{B. Liu \at
              School of Computer Science, Nanjing University of Posts and Telecommunications, and Jiangsu Key Lab of Big Data Security $\&$ Intelligent Processing, Nanjing,
Jiangsu, 210023 China. \\
%              Tel.: +8618951650062\\
%              Fax: +86-25-85866434\\
              \email{bins@ieee.org}
}

\date{Submitted to arXiv on July 23, 2018; Accepted by Knowledge-Based Systems on Jan. 6, 2020}
% The correct dates will be entered by the editor

\maketitle

\begin{abstract}
This paper is concerned with sequential filtering based stochastic optimization (FSO) approaches that leverage a probabilistic perspective to implement the incremental proximity method (IPM).
The present FSO methods are derived based on the Kalman filter (KF) and the
extended KF (EKF). In contrast with typical methods such as stochastic gradient descent (SGD) and IPMs, they do not need to pre-schedule the learning rate for convergence. Nevertheless, they have limitations that inherit from the KF mechanism. As the particle filtering (PF) method outperforms KF and its variants remarkably for nonlinear non-Gaussian sequential filtering problems, it is natural to ask if FSO methods can benefit from PF to get
around of their limitations. We provide an affirmative answer to this
question by developing two PF based stochastic optimizers (PFSOs). For performance evaluation, we apply them to address nonlinear least-square fitting with simulated data, and empirical risk minimization for binary classification of real data sets. Experimental results demonstrate that PFSOs outperform remarkably a benchmark SGD algorithm, the vanilla IPM, and KF-type FSO methods in terms of numerical stability, convergence speed, and flexibility in handling diverse types of loss functions.
\end{abstract}

\section{Introduction}
In this paper, we consider a type of optimization problem that arises in supervised machine learning (ML), formulated as below,
\begin{eqnarray}\label{eqn:obj_fun}
\underset{\mathbf{\theta}\in\mathbb{R}^d}{\min}f(\mathbf{\theta})&=&\underset{\mathbf{\theta}\in\mathbb{R}^d}{\min}\frac{1}{K}\sum_{k=1}^Kf_k(\mathbf{\theta}),\\\nonumber
f_k(\mathbf{\theta})&=&l(\mathbf{\theta}|\mathbf{x}_k,y_k), k=1,\ldots,K, \\\nonumber
\end{eqnarray}
where $\{\mathbf{x}_k,y_k\}_{k=1}^K$ denotes a number $K$ of training data points, $\mathbf{x}\in\mathbb{R}^d$ the feature vector, $y$ the label,
and $l$ the loss function. Such a finite-sum optimization problem is also known as empirical risk minimization (ERM) in the context of ML.
We consider cases in which the number $K$ is so large that it is infeasible to apply any first or second-order optimization methods that need to access gradients of $f_k, k=1,\ldots,K$, per iteration. When $K$ takes a large value, the cost of evaluating and storing such gradients can be too expensive.

The massive scale of modern data sets poses a significant research challenge for developing efficient algorithms to
solve such large-scale ERM problems \cite{reddi2017new,bottou2018optimization}. A large body of work has focused on the SGD methods. In principle, SGD can be regarded as a stochastic version of the gradient descent method \cite{bottou2010large,bottou1998online}. The key idea of SGD is to select a cheap, noisy but unbiased estimate of the true gradient of $f$ for use.
The common practice is to randomly choose one or a subset of the training data points, and then to estimate the gradient based on the chosen data points.
%A basic approach to constitute the subset of training data points is just uniform random sampling with replacement.
%  the training data points used for parameter updating are randomly selected at each iteration, a
SGD requires low computation and memory resources to run, and has a frequently fast initial convergence rate, while its performance is highly dependent on the scheduling of the step-size, also known as the learning rate. One has to elaborately design a schedule of the step-size for one dataset, while this schedule is incapable of adapting to other dataset’s characteristics. To improve the adaptivity or accelerate convergence rate of the vanilla SGD, much work has been conducted such as Momentum \cite{qian1999momentum}, Nesterov accelerated gradient \cite{nesterov1983method}, Adagrad \cite{duchi2011adaptive}, Adadelta \cite{zeiler2012adadelta}, Adaptive Moment Estimation (Adam) \cite{kingma2014adam}, Nesterov-accelerated Adaptive Moment Estimation (Nadam).
These improved methods can perform better than the vanilla SGD, while they also require human effort for hyper-parameter tuning.
% to avoid trapping into local minimums or saddle points. To summarize, there still lacks a generally applicable while theoretically sound approach to let SGD converge fast while globally.

IPM, an alternative approach to SGD, has also raised concerns \cite{parikh2014proximal,bertsekas2011incremental,aky2018the}.
IPM runs a sequence of proximal operators, each of which minimizes a single or a mini-batch of components in the additive cost function. Each proximal operator searches over a constrained local parameter space centered around the solution given in the former iteration. IPM is preferable to SGD for linear cases since it has an analytical solution \cite{parikh2014proximal,bertsekas2011incremental}. For nonlinear cases,
no analytical solution exists anymore; instead, a numerical solver is required at every proximal searching step, rendering the searching process computationally inefficient \cite{parikh2014proximal,bertsekas2011incremental}. Besides, when the estimated minimum approaches the actual minimum, IPM tends to be numerically unstable, due to the lack of a mechanism to reduce the step-size \cite{aky2018the}.

Based on a probabilistic explanation of the proximity operator, KF has recently been applied to improve IPM.
The introduction of the KF iteration into the IPM provides a natural dampening mechanism for parameter updates,
rendering the resulting KF-IPM algorithm much more numerically stable than the vanilla IPM \cite{aky2018the}.
Nevertheless, for KF-IPMs, there are two severe problematic issues that inherit from the intrinsic limitations of KF.
First, KF-IPMs only work for cases in which each component $f_k$ of the cost function $f$ takes a specific form $f_k(\theta)=(y_k-h_k(\theta))^2$; otherwise, the closed-form update formulas are not available. Second, when $h_k$ is highly nonlinear, an extended KF (EKF) should be applied instead of KF; the resulting EKF-IPM algorithm is likely to be divergent, due to model mismatch resulted from the application of the first-order linearization of $h_k$. The presence of the aforementioned problematic issues hinders the KF type IPMs from being widely used in ML tasks. For example, if a logistic loss function $f_k(\theta)\triangleq \log\left(1+\exp\left(-y_k\mathbf{x}_k^{\top}\theta\right)\right)$ is adopted, the KF type IPMs can not be applicable, as the KF updating formula is not available.
Here $A^{\top}$ denotes the transposition of $A$ (we assume that both $\theta$ and $\mathbf{x}$ are column
vectors).

In this paper, we investigate whether stochastic optimization can benefit from the PF theory to
get around the limitations of KF-IPM and EKF-IPM. We provide an affirmative
answer to this research question. Specifically, we develop the PF based stochastic optimization (PFSO) algorithm and show that it owns desirable properties that have been beyond the capability of previous methods. Some related solutions are proposed for ad hoc scenarios, e.g., operations and logistics \cite{gharaei2019joint,gharaei2019integrated,duan2018selective,hoseini2019modelling,gharaei2019modelling,gharaei2019integrated2}. We do not consider such applied methods here since they have no direct connection to the research question of our concern. Instead, we select the most representative and generic algorithms, including the vanilla IPM, KF-IPM, EKF-IPM, and a typical SGD algorithm termed Adaline \cite{bottou1998online}, for performance comparison in Section \ref{sec:experiment}.

The remainder of this paper is organized as follows. In Section \ref{sec:background}, we briefly review the link between KF and
IPM. In Section \ref{sec:pfso}, we present the proposed PFSO method in detail. In Section \ref{sec:discuss},
we discuss connections between our methods and other related work. In Section \ref{sec:experiment},
we present experimental results that demonstrate the superiority of our methods. Finally, we conclude the paper in Section \ref{sec:conclusions}.
\section{Revisit the link between KF and IPM}\label{sec:background}
Here we briefly review the link between KF and IPM. This review provides the necessary background information required for developing the proposed PFSO method in Section \ref{sec:pfso}.

IPMs solve problems of the form (\ref{eqn:obj_fun}) iteratively. Each iteration produces an updated estimate of $\theta$.
Denote the estimate of $\theta$ generated at iteration $k$ as $\theta_k$. Given $\theta_{k-1}$, only a single component $f_k$ of the objective function is involved for generating $\theta_k$. Specifically, $\theta_k$ is obtained through solving the following problem
\begin{equation}\label{eqn:ipm_basic}
\theta_{k}=\operatorname{prox}_{\lambda, f_{k}}\left(\theta_{k-1}\right)=\arg\underset{\theta \in \mathbb{R}^{d}}{\operatorname{min}} \left[f_{k}(\theta)+\lambda\left\|\theta-\theta_{k-1}\right\|_{2, \mathbf{V}^{-1}}^{2}\right],
\end{equation}
where $\operatorname{prox}$ denotes the proximal operator, $\lambda \in \mathbb{R}_{+}$ a regularization parameter, $\mathbf{V}\in\mathbb{R}^{d\times d}$ a symmetric positive definite matrix,
$\left\|\mathbf{a}\right\|_{2,\mathbf{V}^{-1}}^2\triangleq \mathbf{a}^{\top}\mathbf{V}^{-1}\mathbf{a}$, $\mathbf{V}^{-1}$ the
inverse of $\mathbf{V}$. The same as in \cite{aky2018the}, we slightly abuse the notation $f_k$ here for simplicity, which stands for $f_{j_k}$,
where $j_k$ is a random sample drawn uniformly from the set $\{1,2,\ldots,n\}$.
For cases in which $f_{k}(\theta)=\left(y_{k}-\mathbf{x}_{k}^{\top} \theta\right)^{2}$, the solution to (\ref{eqn:ipm_basic})
is shown to be \cite{aky2018the}:
\begin{equation}\label{eq:ipm}
\theta_{k}=\theta_{k-1}+\frac{\mathbf{V} \mathbf{x}_{k}\left(y_{k}-\mathbf{x}_{k}^{\top} \theta_{k-1}\right)}{\lambda+\mathbf{x}_{k}^{\top} \mathbf{V} \mathbf{x}_{k}}
\end{equation}
Now we define a model that consists of a prior density function and a likelihood function as follows,
\begin{equation}\label{eq:model_kf_ipm}
p(\theta)=\mathcal{N}\left(\theta;\theta_{0}, \mathbf{V}_{0}\right), \quad p\left(y_{k} | \theta\right)=\mathcal{N}\left(y_{k};\mathbf{x}_{k}^{\top} \theta, \lambda\right),
\end{equation}
where $\mathcal{N}\left(\theta; \mathbf{A},\mathbf{B}\right)$ denotes a Gaussian distribution with mean vector $\mathbf{A}$ and covariance matrix $\mathbf{B}$.
Let $y_{1:k}=\{y_1,\ldots,y_k\}$. Since both the prior density and the likelihood function are Gaussian, the posterior distribution
$p(\theta|y_{1:k})$ is also Gaussian. Let $p(\theta|y_{1:k})=\mathcal{N}\left(\theta;\theta_{k}, \mathbf{V}_{k}\right)$.
Then we have \cite{haykin2001kalman,aky2018the}:
\begin{equation}\label{eq:kf-ipm}
\theta_{k}=\theta_{k-1}+\frac{\mathbf{V}_{k-1} \mathbf{x}_{k}\left(y_{k}-\mathbf{x}_{k}^{\top} \theta_{k-1}\right)}{\lambda+\mathbf{x}_{k}^{\top} \mathbf{V}_{k-1} \mathbf{x}_{k}} ,
\end{equation}
\begin{equation}\label{eq:V_kf-ipm}
\mathbf{V}_{k}=\mathbf{V}_{k-1}-\frac{\mathbf{V}_{k-1} \mathbf{x}_{k} \mathbf{x}_{k}^{\top} \mathbf{V}_{k-1}}{\lambda+\mathbf{x}_{k}^{\top} \mathbf{V}_{k-1} \mathbf{x}_{k}},
\end{equation}
The above equations (\ref{eq:kf-ipm})-(\ref{eq:V_kf-ipm}) constitute the recursion of KF-IPM at iteration $k$. It is shown that the term $\mathbf{V}$ in (\ref{eq:ipm}) is now replaced with $\mathbf{V}_{k-1}$ in (\ref{eq:kf-ipm}). The corresponding proximity operator addressed by KF-IPM can be formulated as
\begin{equation}\label{eqn:proximal_operator_kf_ipm}
\theta_{k}=\operatorname{prox}_{\lambda, f_{k}}\left(\theta_{k-1}\right)=\arg\underset{\theta \in \mathbb{R}^{d}}{\operatorname{min}}\left[f_{k}(\theta)+\lambda\left\|\theta-\theta_{k-1}\right\|_{2, \mathbf{V}_{k-1}^{-1}}^{2}\right].
\end{equation}
Comparing (\ref{eqn:proximal_operator_kf_ipm}) with (\ref{eqn:ipm_basic}), one can see that KF-IPM is a special type of IPMs that tunes its matrix parameter $\mathbf{V}$ across iterations.
It has been demonstrated that, by adapting $\mathbf{V}$ according to (\ref{eq:V_kf-ipm}),
KF-IPM is markedly more numerically stable than the vanilla IPM \cite{aky2018the}.

The above KF-IPM recursion is only applicable for cases in which $y_k\approx \mathbf{x}_k^{\top}\theta$ and $f_k$ takes the form $f_k=(y_k-\mathbf{x}_k^{\top}\theta)^2$, $k=1,\ldots,K$.
The EKF-IPM is derived to handle cases in which $y_k\approx h_k(\theta)$, $f_k$ takes the form $f_k=(y_k-h_k(\theta))^2$, $k=1,\ldots,K$, and $h_k$ is nonlinear and differentiable.
The recursion of EKF-IPM is \cite{aky2018the}
\begin{equation}\label{eq:ekf-ipm}
\theta_k=\theta_{k-1}+\frac{\mathbf{V}_{k-1}\mathbf{a}_k(y_k-h_k(\theta_{k-1}))}{\lambda+\mathbf{a}_k^{\top}\mathbf{V}_{k-1}\mathbf{a}_k},
\end{equation}
\begin{equation}\label{eq:V_ekf-ipm}
\mathbf{V}_{k}=\mathbf{V}_{k-1}-\frac{\mathbf{V}_{k-1}\mathbf{a}_k\mathbf{a}_k^{\top}\mathbf{V}_{k-1}}{\lambda+\mathbf{a}_k^{\top}\mathbf{V}_{k-1}\mathbf{a}_k},
\end{equation}
where $\mathbf{a}_k=\triangledown_{\theta}h_k(\theta_{k-1})$, $\triangledown_{\theta}$ denotes the gradient operator with respect to $\theta$.

Despite advantages over the vanilla IPM, such KF-IPMs have two unresolved issues that
keep them from being more widely applied. The first one is that to apply KF-IPMs, the component function $f_k$ should take a specific form $f_k(\theta)=(y_k-h_k(\theta))^2$. ML applications often involve other forms of nonlinear loss functions, e.g.,
the logistic loss $f_k(\theta)\triangleq \log\left(1+\exp\left(-y_k\mathbf{x}_k^{\top}\theta\right)\right)$.
For such cases, neither KF-IPM nor EKF-IPM can be employed.
The second issue is that, if $h_k$ is highly nonlinear, the estimate given by EKF-IPM is likely to be divergent
due to a significant model mismatch caused by the first-order linearization of $h_k$.
In the next section, we propose PFSO methods that elegantly resolve the above issues.
\section{The Proposed Particle-based Stochastic Optimization Methods}\label{sec:pfso}
In this section, we show how to apply PF as an alternative of KF to address the stochastic optimization problem formulated in (\ref{eqn:obj_fun}). We first present a generic particle-based scheme that extends KF-IPMs.
Then we introduce the proposed algorithms to implement this scheme.
\subsection{A Particle-based Scheme for Stochastic Optimization}
Let consider a model defined by a prior density function and a likelihood function as follows,
\begin{equation}\label{eq:model_pf_ipm}
p(\theta)=\mathcal{N}\left(\theta;\theta_{0}; \mathbf{V}_{0}\right); \quad p\left(y_{k} | \theta\right)= \exp(-f_k(\theta)/\lambda).
\end{equation}
In this model, the prior density function is the same as that in (\ref{eq:model_kf_ipm}), while the form of the likelihood function is different.
Recall that the model defined by (\ref{eq:model_kf_ipm}) is the basis on which KF-IPM is derived.
At iteration $k$, KF-IPM outputs $\theta_k$ as an updated estimate of $\theta$, which is exactly the mean of the posterior $\pi_k(\theta)\triangleq p(\theta|y_{1:k})$.
In model (\ref{eq:model_pf_ipm}) considered here, we allow $f_k$ to take any form,
while, for ease in presentation, we focus on the logistic loss $f_k(\theta)=\log\left(1+\exp\left(-y_k\mathbf{x}_k^{\top}\theta\right)\right)$ in what follows.
For this logistic loss case, neither KF-IPM nor EKF-IPM is applicable for calculating the mean of the posterior,
due to the lack of analytical recursion equations.
Here we resort to particle-based methods to estimate the mean $\theta_k$ of the posterior $\pi_k(\theta)$.

The basic idea is to run a sequential importance sampling (SIS) procedure to simulate
$\pi_k(\theta), k=1,2,\ldots,K$. Then one can seek an estimate $\hat{\theta}_k$ of $\theta_k$
by making use of the samples yielded from the simulation. The SIS procedure is the backbone of all PF methods.
Following the standard in PF related literature, we call sampled values of $\theta$ as ``particles" in what follows.
Suppose that, standing at the beginning of iteration $k$, we have a weighted particle set $\{\theta_{k-1}^i,\omega_{k-1}^i\}_{i=1}^N$,
which provides a Monte Carlo approximation to $\pi_{k-1}(\theta)$, namely
\begin{equation}
\pi_{k-1}(\theta)\simeq\sum_{i=1}^N\omega_{k-1}^i\delta_{\theta_{k-1}^i},
\end{equation}
where $\delta_x$ denotes the delta-mass function located at $x$. Then a two-stage operation is performed in the SIS procedure.
First, draw a set of new particles $\{\theta_{k}^i\}_{i=1}^N$ from a proposal function $q_k$: $\theta_{k}^i\sim q_k(\cdot)$.
Then, calculate the importance weights of these particles as follows
\begin{eqnarray}\label{eqn:SIS_weight}
\hat{\omega}_{k}^i&=&\omega_{k-1}^i\times\frac{\pi_k(\theta_{k}^i)}{q_k(\theta_{k}^i)}, i=1,\ldots,N,\\
\omega_{k}^i&=&\frac{\hat{\omega}_{k}^i}{\sum_{j=1}^N\hat{\omega}_{k}^j}, i=1,\ldots,N.
\end{eqnarray}
Under mild conditions and with an appropriate design of the proposal function,
this updated particle set can provide a satisfactory Monte Carlo approximation
to $\pi_{k}(\theta)$ \cite{arulampalam2002tutorial}, namely,
\begin{equation}
\pi_{k}(\theta)\simeq\sum_{i=1}^N\omega_{k}^i\delta_{\theta_{k}^i}.
\end{equation}
Then one can calculate $\hat{\theta}_k$ as below
\begin{equation}
\hat{\theta}_k=\frac{1}{N}\sum_{i=1}^N\omega_{k}^i\theta_{k}^i.
\end{equation}

The SIS algorithm has a seriously problematic issue, namely the variance of the importance weights increases stochastically over
iterations \cite{arulampalam2002tutorial}.
The variance increase will cause the phenomenon of particle degeneracy, which means that, after a few iterations, one of the normalized
importance weights approaches one, while others tend to zero. To reduce particle degeneracy, a resampling procedure is usually
used to eliminate particles with low importance weights and duplicate particles with high importance weights \cite{arulampalam2002tutorial}.
Several resampling schemes, such as residual resampling and minimum variance sampling, have been proposed,
while their impacts on the final performance are not significantly different among each other \cite{van2001unscented}.
We used residual resampling in all our experiments in Section \ref{sec:experiment}.
A pseudo-code illustrating the PF scheme for stochastic optimization is presented in Algorithm \ref{algo:SIS}.
\begin{algorithm}[!htb]
\caption{PFSO: A generic PF scheme for stochastic optimization}
\label{algo:SIS}
\begin{algorithmic}[1]
\STATE Initialization: Draw a random sample $\{\theta_0^i\}_{i=1}^N$ from $\pi_0(\theta_0)$. Set $\omega_0^i=1/N$, $\forall i$. (Here and in what follows, `$\forall i$' means `for all $i$ in $\{1,\ldots,N\})$'\;
\FOR{$k=1,\ldots, K$ }
\STATE Sample $\theta_{k}^i\sim q_k(\theta_k)$, $\forall i$;
\STATE Calculate the importance weights of the particles: $\hat{\omega}_{k}^i=\frac{\pi_k(\theta_{k}^i)}{q_k(\theta_{k}^i)}$,
    $\forall i$;
\STATE Normalize the importance weights:
    $\omega_{k}^i=\frac{\hat{\omega}_{k}^i}{\sum_{i=1}^N\hat{\omega}_{k}^i}$, $\forall i$;
\STATE Set $\hat{\theta}_k=\sum_{i=1}^N\omega_{k}^i\theta_{k}^i$;
\STATE Resampling step: Eliminate/duplicate samples with low/high importance weights, respectively,
yielding $N$ equally weighted particles $\theta_k^i$ approximately distributed as $\pi_k(\theta)$; Set $\omega_k^i=1/N, \forall i$.
\ENDFOR
\end{algorithmic}
\end{algorithm}

The PFSO scheme provides the basis for developing PF based stochastic optimization algorithms,
while a critical issue, namely the choice of the proposal function $q_k$, has not been addressed so far.
For sampling from $\pi_k(\theta)$, we hope that, by choosing an appropriate $q_k$,
the variance of the importance weights can tend to zero. The choice of the proposal function plays an important role in
reducing the variance of the importance weights \cite{van2001unscented,liu2008particle,arulampalam2002tutorial}. An empirical guideline
is to choose one that mimics the target distribution but has heavier tails. In the following subsections,
we present two algorithm designs to implement the PFSO scheme.
\subsection{Kernel Smoothing based PFSO}
Here we adopt a kernel smoothing technique,
referred to as Liu and West method \cite{liu2001combined} in the literature,
to generate new particles in the context of PFSO. The resulting algorithm is termed kernel smoothing based PFSO (KS-PFSO).
Suppose that, standing at the beginning of iteration $k$, we have at hand a
weighted particle set $\{\theta_{k-1}^i,\omega_{k-1}^i\}_{i=1}^N$, that
satisfies $\pi_{k-1}(\theta_{k-1})\simeq\sum_{i=1}^N\omega_{k-1}^i\theta_{k-1}^i$.
We calculate the mean and variance of $\pi_{k-1}(\theta_{k-1})$ with particle approximation.
Denote the approximated mean and
variance by $\hat{\mathbf{m}}_{k-1}$ and $\hat{\mathbf{V}}_{k-1}$, respectively.
Then sample $\epsilon^i\thicksim\mathcal{N}(\textbf{0}_d,\gamma \hat{\mathbf{V}}_{k-1})$ and set
\begin{equation}\label{eqn:kde_sample}
\theta_k^i=\rho\theta_{k-1}^i+(1-\rho)\mathbf{m}_{k-1}+\epsilon^i,
\end{equation}
where $\textbf{0}_d$ denotes a $d$-dimensional zero valued vector, $\rho$ and $\gamma$ are free parameters chosen to satisfy $\rho^2+\gamma=1$, ensuring the mean and variance of these new-born particles to be correct \cite{liu2001combined}.
The operation in (\ref{eqn:kde_sample}) brings two desirable effects. First, particle rejuvenation is achieved. Second,
it retains the mean of the particles, and meanwhile avoids over-dispersion of these new particles \cite{liu2001combined}.
Set the value of $\rho$ to approach 1, then, from (\ref{eqn:kde_sample}), one can infer that the position of $\theta_k^i$ will be close to $\theta_{k-1}^i$, $i=1,\ldots,N$ and the corresponding proposal $q_k(\theta)$ from which the new particles are drawn is approximately equivalent to $\pi_{k-1}(\theta)$.
According to (\ref{eqn:SIS_weight}), the importance weights of the new-born particles are obtained as follows
\begin{eqnarray}\label{eqn:ks_pfso_weight}
\hat{\omega}_{k}^i&=&\omega_{k-1}^i p\left(y_{k} | \theta_{k}^i\right), i=1,\ldots,N,\\
\omega_{k}^i&=&\frac{\hat{\omega}_{k}^i}{\sum_{j=1}^N\hat{\omega}_{k}^j}, i=1,\ldots,N.
\end{eqnarray}
Note that, after the resampling step, all particles have the same importance weight $1/N$, and thus (\ref{eqn:ks_pfso_weight}) can be substituted with a simpler calculation, namely $\hat{\omega}_{k}^i=p\left(y_{k} | \theta_{k}^i\right)$.
The KS-PFSO algorithm is summarized as follows in Algorithm \ref{algo:ks-pfso}.
\begin{algorithm}[!htb]
\caption{The KS-PFSO Algorithm}
\label{algo:ks-pfso}
\begin{algorithmic}[1]
\STATE Initialization: Draw random particles $\{\theta_0^i\}_{i=1}^N$ from $\pi_0(\theta_0)\triangleq \mathcal{N}(\theta_0,\hat{V}_0)$.
Set $\hat{\mathbf{m}}_0=\theta_0$ and $\omega_0^i=1/N$, $\forall i$.
\FOR{$k=1,\ldots, K$}
\STATE Sample $\theta_{k}^i$ using Eqn.(\ref{eqn:kde_sample}), $\forall i$.
\STATE Calculate importance weights of the particles: $\hat{\omega}_{k}^i=p(y_k|\theta_k^i)$,
    $\forall i$.
\STATE Normalize the weights:
    $\omega_{k}^i=\frac{\hat{\omega}_{k}^i}{\sum_{i=1}^N\hat{\omega}_{k}^i}$, $\forall i$.
\STATE Set $\hat{\theta}_k=\sum_{i=1}^N\omega_{k}^i\theta_{k}^i$.
\STATE Set $\hat{\mathbf{m}}_{k}=\hat{\theta}_k$,
$\hat{\mathbf{V}}_{k}=\sum_{i=1}^N\omega_{k}^i(\theta_{k}^i-\hat{\mathbf{m}}_{k})(\theta_{k}^i-\hat{\mathbf{m}}_{k})^T$.
\STATE Resampling step: the same as in Algorithm \ref{algo:SIS}.
\ENDFOR
\end{algorithmic}
\end{algorithm}
\subsection{Random Perturbation Assisted PFSO}
In KS-PFSO, as shown in Algorithm \ref{algo:ks-pfso}, a resampling procedure is adopted to reduce particle degeneracy.
This procedure results in multiple copies of the fittest particles and removal of low weight particles,
which may lead to a phenomenon called particle impoverishment \cite{arulampalam2002tutorial}.
The extreme case is that, after the resampling step, all $N$ particles take the identical value.
The effect of particle impoverishment can be more disastrous if $N$ takes a small value. Here we present a random perturbation assisted PFSO (RP-PFSO), in which a random perturbation step
is introduced to strengthen particle diversity after the resampling step.
The idea is to move each particle, say $\theta_k^i$, to a new state $\tilde{\theta}_{k}^{i}$, if the following condition satisfies
\begin{equation}\label{eqn:rp}
v\leq\min\left\{1,\frac{p(y_k|\tilde{\theta}_{k}^{i})}{p(y_k|\theta_{k}^i)}\right\},
\end{equation}
where $v\thicksim\mathcal{U}_{[0,1]}$ and $\mathcal{U}_{[0,1]}$ denotes uniform distribution over [0,1].
The candidate state $\tilde{\theta}_{k}^{i}$ is a random sample drawn from a Gaussian distribution centered at $\theta_k^i$.
As shown in (\ref{eqn:rp}), as long as $p(y_k|\tilde{\theta}_{k}^{i})\geq p(y_k|\theta_{k}^i)$, the new state $\tilde{\theta}_{k}^{i}$ will be accepted. Even if $p(y_k|\tilde{\theta}_{k}^{i})< p(y_k|\theta_{k}^i)$, the new state $\tilde{\theta}_{k}^{i}$ will also be accepted with a probability $p(y_k|\tilde{\theta}_{k}^{i})/p(y_k|\theta_{k}^i)$. Therefore, we improve the particle diversity by performing a local exploration biased toward higher likelihood regions recommended by $y_k$.
The pseudo-code of the RP-PFSO algorithm is presented in Algorithm \ref{algo:mcmc-pfso}.
Empirical results in Section \ref{sec:experiment} show that RP-PFSO is preferable to KS-PFSO
when the particle size $N$ takes a small value, corresponding to cases wherein the posterior distribution is under-sampled.
\begin{algorithm}[!htb]
\caption{The RP-PFSO Algorithm}
\label{algo:mcmc-pfso}
\begin{algorithmic}[1]
\STATE Initialization: the same as in Algorithm \ref{algo:ks-pfso}\;
\FOR{$k=1,\ldots, K$ }
\STATE Sample $\theta_{k}^i$ according to (\ref{eqn:kde_sample}), $\forall i$;
\STATE Calculate the importance weights: $\hat{\omega}_{k}^i=p(y_k|\theta_k^i)$,
    $\forall i$;
\STATE Normalize the importance weights: $\omega_{k}^i=\frac{\hat{\omega}_{k}^i}{\sum_{i=1}^N\hat{\omega}_{k}^i}$, $\forall i$;
\STATE Resampling step: the same as in Algorithm \ref{algo:SIS};
\STATE Random perturbation step: for $\forall i$, perform the following three operations:
\begin{itemize}
\item Sample $v\thicksim\mathcal{U}_{[0,1]}$;
\item Sample a new particle $\tilde{\theta}_{k}^{i}$ from a Gaussian distribution centered at $\theta_k^i$;
\item Set $\theta_k^i=\tilde{\theta}_{k}^i$ if (\ref{eqn:rp}) holds.
\end{itemize}
\STATE Set $\hat{\theta}_k=\sum_{i=1}^N\omega_{k}^i\theta_{k}^i$.
\STATE Set $\hat{\mathbf{m}}_{k}=\hat{\theta}_k$,
$\hat{\mathbf{V}}_{k}=\sum_{i=1}^N\omega_{k}^i(\theta_{k}^i-\hat{\mathbf{m}}_{k})(\theta_{k}^i-\hat{\mathbf{m}}_{k})^T$.
\ENDFOR
\end{algorithmic}
\end{algorithm}
\section{Connections to Related Algorithms}\label{sec:discuss}
Here we discuss connections between our methods and the major related works in the literature.
\subsection{Connections to PF methods for dynamic state filtering}
Most of PF algorithms are developed in the context of dynamic state filtering \cite{gordon1993novel,arulampalam2002tutorial}.
In these methods, a state transition prior is usually precisely defined and then adopted as the proposal to generate new particles.
For PFSOs, there is no state transition prior function defined, and the new particles are sampled by using a kernel smoothing technique, termed Liu and West method \cite{liu2001combined}. In PFSOs, the proposal distribution, from which new born particles
are sampled, is an approximation of the posterior yielded from the last iteration.
Besides, RP-PFSO is related with the improved PF methods presented in \cite{liu2008single,gilks2001following},
which adopt a Markov Chain Monte Carlo moving step to strengthen particle diversity.

The fundamental difference between the presented PFSO approach and the conventional PF methods developed for dynamic state filtering can be summarized as follows. The former treats the variable to be estimated, namely $\theta$ here, as an unknown static parameter; while, the latter models it as a dynamic state that changes over time.
\subsection{Connections to PF methods for static parameter estimation}
The PFSOs proposed here have connections to existent PF methods derived for static model parameter estimation,
the most representative of which are those presented in \cite{chopin2002sequential,ridgeway2003sequential}.
Specifically, both RP-PFSO and methods of \cite{chopin2002sequential,ridgeway2003sequential} use a random perturbation step to
bypass particle impoverishment after the resampling step. The difference lies in that, the former employs Liu and West method to
 generate new particles besides the random perturbation step, while the latter relies completely on an MCMC moving step to
 generate new particles.
Besides, in the context of PFSOs, the index of the data point that arrives at iteration $k$ is $j_k$,
 which is randomly and uniformly drawn from the index set $\{1,\ldots,K\}$.
 For methods in \cite{chopin2002sequential,ridgeway2003sequential}, the data point processed at iteration $k$
 is exactly the $k$th item of the training data set.
\subsection{Connections to filtering methods for optimization}
As a type of recursive filter based stochastic optimization methods, our PFSO algorithm has connections to all KF-type stochastic optimization methods, e.g., in \cite{patel2016kalman,aky2018the,bell1993iterated,bertsekas1996incremental,ho1962stochastic}.
In this paper, we substitute KFs with specially designed PFs in the context of stochastic optimization.
With aid of PF, our methods allow any form of the loss function to be employed,
while the KF-type method requires an ad hoc form of the loss function.
Besides, with PF as its backbone, the presented PFSO algorithm is well suitable for handling highly nonlinear models that bother
KF-type methods.

Our PFSO methods also find connections to several PF based global optimization (PFO) methods
in \cite{stinis2012stochastic,liu2017posterior,liu2016particle}, as
they all fall within an algorithmic framework termed Sequential Monte Carlo (SMC) sampler \cite{del2006sequential}.
In this framework, the task of optimizing an objective function $f(\theta)$ is translated to sampling a
sequence of target distributions $\pi_k(\theta)$, and then evaluating the optimum based on the samples. A basic difference
between the existent PFO approaches and the PFSO here is as follows. To evaluate a candidate value of $\theta$ in searching the minimum of (\ref{eqn:obj_fun}), the former
needs a full computation of the objective function $f(\theta)$, while the latter only needs to compute a component function value, say $f_k(\theta)$. When the number of the additive components $K$ is large, the former can be inapplicable, while the latter can still work elegantly.
\section{Experimental Evaluations}\label{sec:experiment}
We seek to experimentally validate two claims about our methods.
The first is that the proposed PFSO methods outperform both the KF-type IPMs and the vanilla IPM in handling nonlinear models in the context of stochastic optimization.
We tested this claim across two synthetic nonlinear least-square fitting cases and 6 real-world binary classification applications.
The second claim is that our PFSO algorithms own the flexibility for handling diverse types of loss functions with a performance guarantee.
We tested this claim with 6 real-world binary classification applications.
All objective functions involved in our experiments are nonlinear.
\subsection{Synthetic data experiments}\label{sec:toy_data}
We considered two nonlinear model fitting experiments, in which EKF-IPM, UKF-IPM and the vanilla IPM are involved for
performance comparison. The term UKF-IPM is the abbreviation of unscented Kalman filter (UKF) based IPM. It is obtained by
substituting the EKF recursion, namely (\ref{eq:ekf-ipm})-(\ref{eq:V_ekf-ipm}), with a UKF recursion.
See \cite{julier2004unscented} for details about UKF.
\subsubsection{Least-square fitting of a sigmoid function}\label{sec:sigmoid}
The setting of this experiment is borrowed from \cite{aky2018the}.
The cost function to be minimized takes the form (\ref{eqn:obj_fun}) with $f_k$ and $h_k$ defined as follows
\begin{eqnarray}
f_k(\theta)&=&(y_k-h_k(\theta))^2,\\
h_k(\theta)&=&\frac{1}{1+\exp(-\alpha-\beta^T\mathbf{x}_k)},
\end{eqnarray}
where $\mathbf{x}_k\in\mathbb{R}^{d-1}$, $\alpha\in\mathbb{R}$, $\beta\in\mathbb{R}^{d-1}$, $\theta\triangleq[\alpha,\beta^T]^T$,
$y_k\in\mathbb{R}$. We simulated $K=3000$ training data points $\{\mathbf{x}_k,y_k\}_{k=1}^K$ for use, in which
$\mathbf{x}_k\sim \mathcal{N}(\textbf{0}_{d-1},\textbf{I}_{d-1})$
and, given $\mathbf{x}_k$, $y_k$ is set to be $\frac{1}{1+\exp(-\alpha-\beta^T\mathbf{x}_k)}+\sqrt{\lambda}n_k$, where $n_k\sim \mathcal{N}(0,1)$,
$\textbf{I}_d$ denotes a $d$ by $d$ identity matrix.
In this experiment, we set $d$ and $\lambda$ at 2 and 0.1, respectively.
The initial estimate $\theta_0$ of $\theta$ for IPM is randomly selected from a uniform distribution that centered around $\theta$.
As the model is nonlinear, the vanilla IPM applied here is actually an approximate nonlinear IPM,
which applies an iterative numerical solver at each iteration.
The initial particle values of the PFSO methods are set to be $\theta_0$.
The EKF-IPM and UKF-IPM are initialized with $(\theta_0, \mathbf{V}_0)$, where $\mathbf{V}_0=\textbf{I}_d$.
As all methods adopt the same initial estimate of $\theta$, a fair
comparison of these methods is thus guaranteed. The particle size $N$ of PFSOs is set at $500$.
Given an estimate $\theta_k$ of $\theta$ yielded at iteration $k$, the corresponding cost or loss value $C(k)$ is defined to be
$C(k)=\frac{1}{K}\sum_{i=1}^Kf_i(\theta_k)$. A normalized cost (NC) is adopted as the performance metric, defined as $NC(k)=C(k)/C(0)$.

\begin{figure}[!htb]
\begin{tabular}{c}
\centerline{\includegraphics[width=3.8in,height=2.5in]{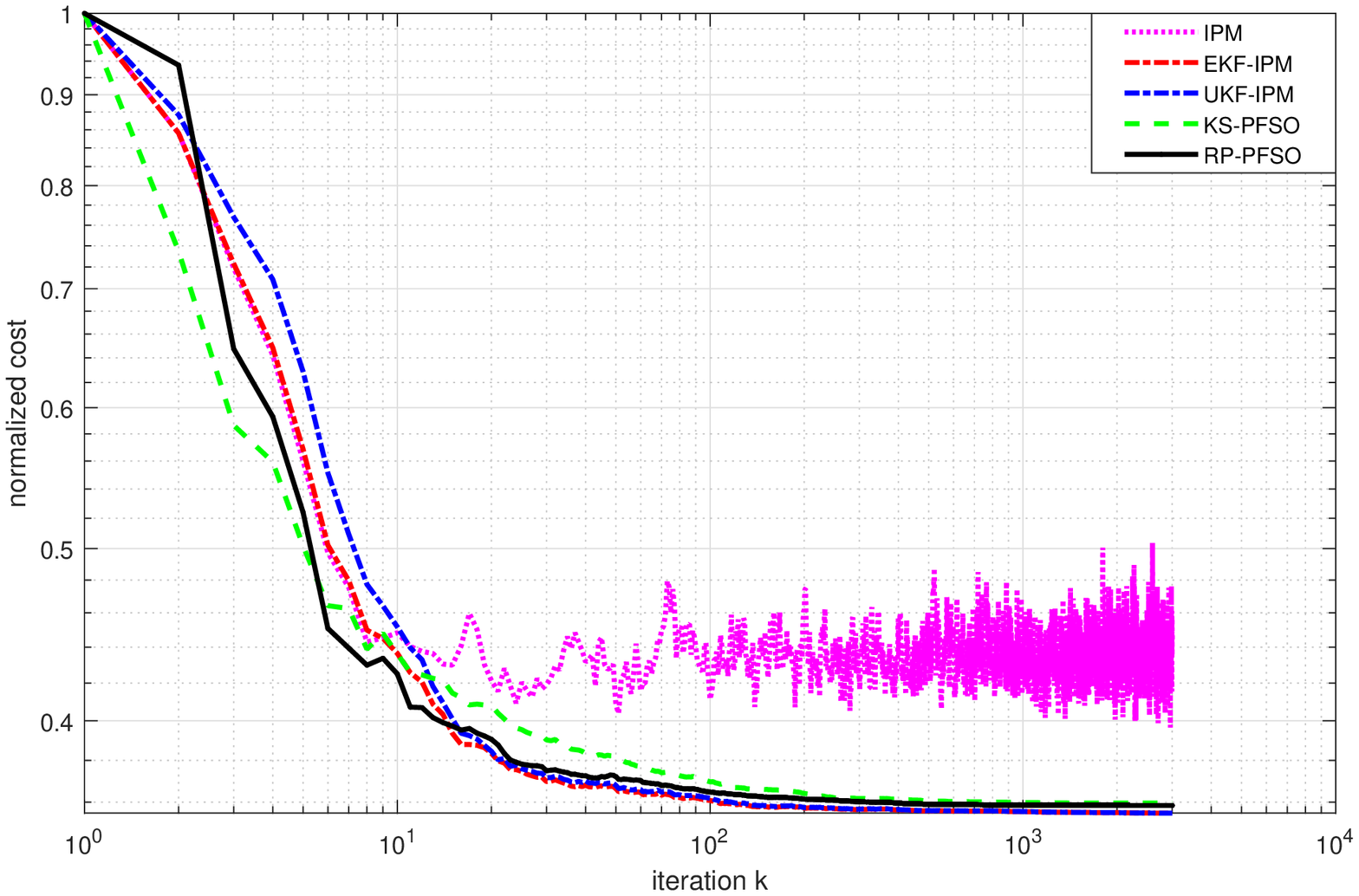}}\\
\centerline{\includegraphics[width=3.8in,height=2.5in]{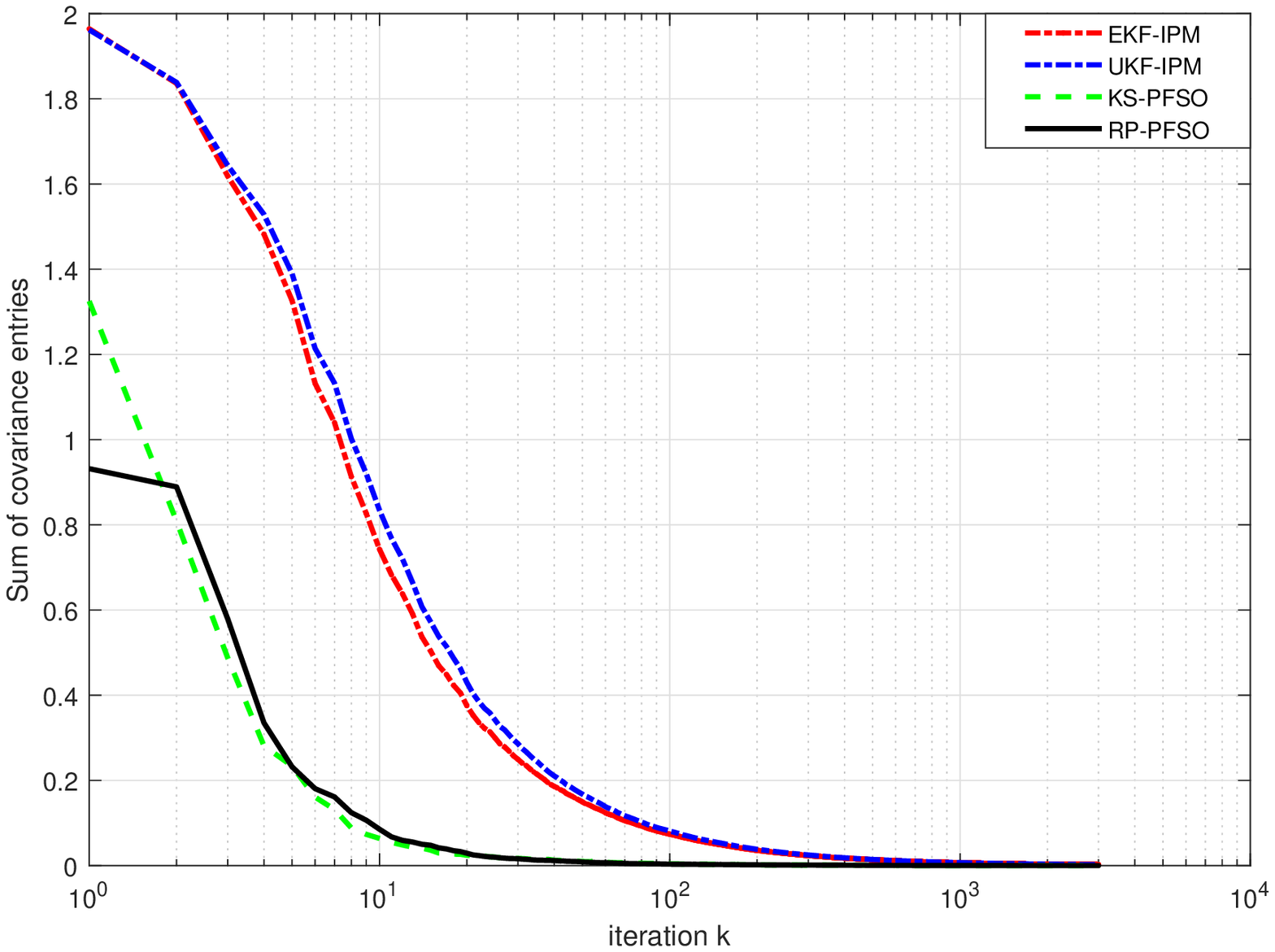}}
\end{tabular}
\caption{Fitting a sigmoid function using the vanilla IPM, EKF-IPM, UKF-IPM and the proposed PFSOs.
Each algorithm is run 30 times independently. This figure shows the result averaged over these runs.
The upper panel shows the evolution of the normalized cost over the iterations. One can see that
KS-PFSO and RP-PFSO perform comparatively with EKF-IPM and UKF-IPM, and significantly better than
the vanilla IPM. The vanilla IPM suffers from instability in iterations of the final phase.
The lower panel depicts the sum of the absolute values of the entries of $\hat{\mathbf{V}}$ per iteration, which implies that
KS-PFSO and RP-PFSO converge faster than the other methods.} \label{fig:simu_case}
\end{figure}

\begin{figure}[!htb]
\begin{tabular}{c}
\centerline{\includegraphics[width=3.8in,height=2.6in]{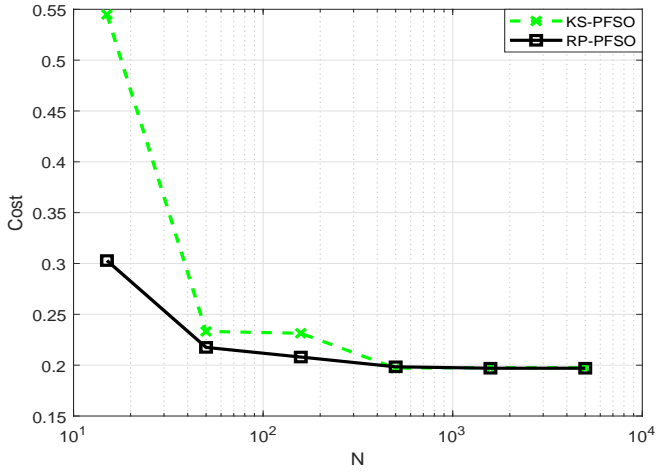}}
\end{tabular}
\caption{The particle size $N$ of the PFSOs vs. the achieved minimum cost. This result is obtained for the experimental case
presented in Subsection \ref{sec:sigmoid}.} \label{fig:simu_N_vs_error}
\end{figure}

Each algorithm is run 30 times independently and an averaged $NC(k)$, $k=0,\ldots,K$, is calculated over these runs.
The experimental result is presented in Fig.\ref{fig:simu_case} and Tables \ref{Table:top_fig1}-\ref{Table:bottom_fig1}.
As is shown, KS-PFSO and RP-PFSO have a faster convergence rate during the first 10 iterations. They perform comparatively with EKF-IPM and UKF-IPM in terms of the final cost.
Both the KF-type and the particle-based methods are numerically stable, while the
vanilla IPM suffers from instability in the final stage.
In Fig.\ref{fig:simu_N_vs_error} and Table \ref{Table:fig2}, the experimental result on the relationship between the PFSO performance and the particle size is presented. It shows that, when $N$ achieves 500, the performance of the particle-based methods tends to be stable. Besides, RP-PFSO performs better than KS-PFSO when $N$ takes small values, which represent cases in which the posterior is under-sampled.
\begin{table*}[!phtb]\caption{A tabular show for the top panel of Fig.\ref{fig:simu_case}}\label{Table:top_fig1}
\centering
\begin{tabular}{cccccccc }
    \hline $k$ & 1 & 4    & 7    &  10    & 100 & 1000 & 3000\\
    \hline IPM & 1 &0.642 &0.476 & 0.448 & 0.412 & 0.444 & 0.441 \\
\hline EKF-IPM & 1 &0.648 &0.480 & 0.437 & 0.361 & 0.356 & 0.355\\
\hline UKF-IPM & 1 &0.709 &0.510 & 0.452 & 0.362 & 0.356 & 0.355\\
\hline KS-PFSO & 1 &0.559 &0.462 & 0.437 & 0.370 & 0.360 & 0.359\\
\hline RP-PFSO & 1 &0.593 &0.440 & 0.425 & 0.365 & 0.359 & 0.358\\
\hline
\end{tabular}
\end{table*}
\begin{table*}[!phtb]\caption{A tabular show for the bottom panel of Fig.\ref{fig:simu_case}}\label{Table:bottom_fig1}
\centering
\begin{tabular}{cccccccc }
    \hline $k$ & 1     & 4    & 7    &  10    & 100   & 1000  & 3000\\
\hline EKF-IPM & 1.964 &1.483 &1.040 & 0.742  & 0.074 & 0.007 & 0.002\\
\hline UKF-IPM & 1.962 &1.528 &1.133 & 0.838  & 0.081 & 0.007 & 0.002\\
\hline KS-PFSO & 1.325 &0.281 &0.130 & 0.064  & 0.005 & 0.0002& 0\\
\hline RP-PFSO & 0.932 &0.337 &0.162 & 0.085  & 0.004 & 0.0002& 0\\
\hline
\end{tabular}
\end{table*}
\begin{table*}[!phtb]\caption{A tabular show for Fig.\ref{fig:simu_N_vs_error}}\label{Table:fig2}
\centering
\begin{tabular}{ccccccc}
    \hline $N$ & 15     & 50    & 158    &  500    & 1580   & 5000 \\
\hline KS-PFSO & 0.545  &0.233  &0.231   & 0.198   & 0.197  & 0.197 \\
\hline RP-PFSO & 0.303  &0.217  &0.208   & 0.198   & 0.197  & 0.197 \\
\hline
\end{tabular}
\end{table*}
\subsubsection{Least-square fitting of a highly nonlinear model}\label{sec:park91a}
\begin{figure}[!htb]
\begin{tabular}{c}
\centerline{\includegraphics[width=3.8in,height=2.5in]{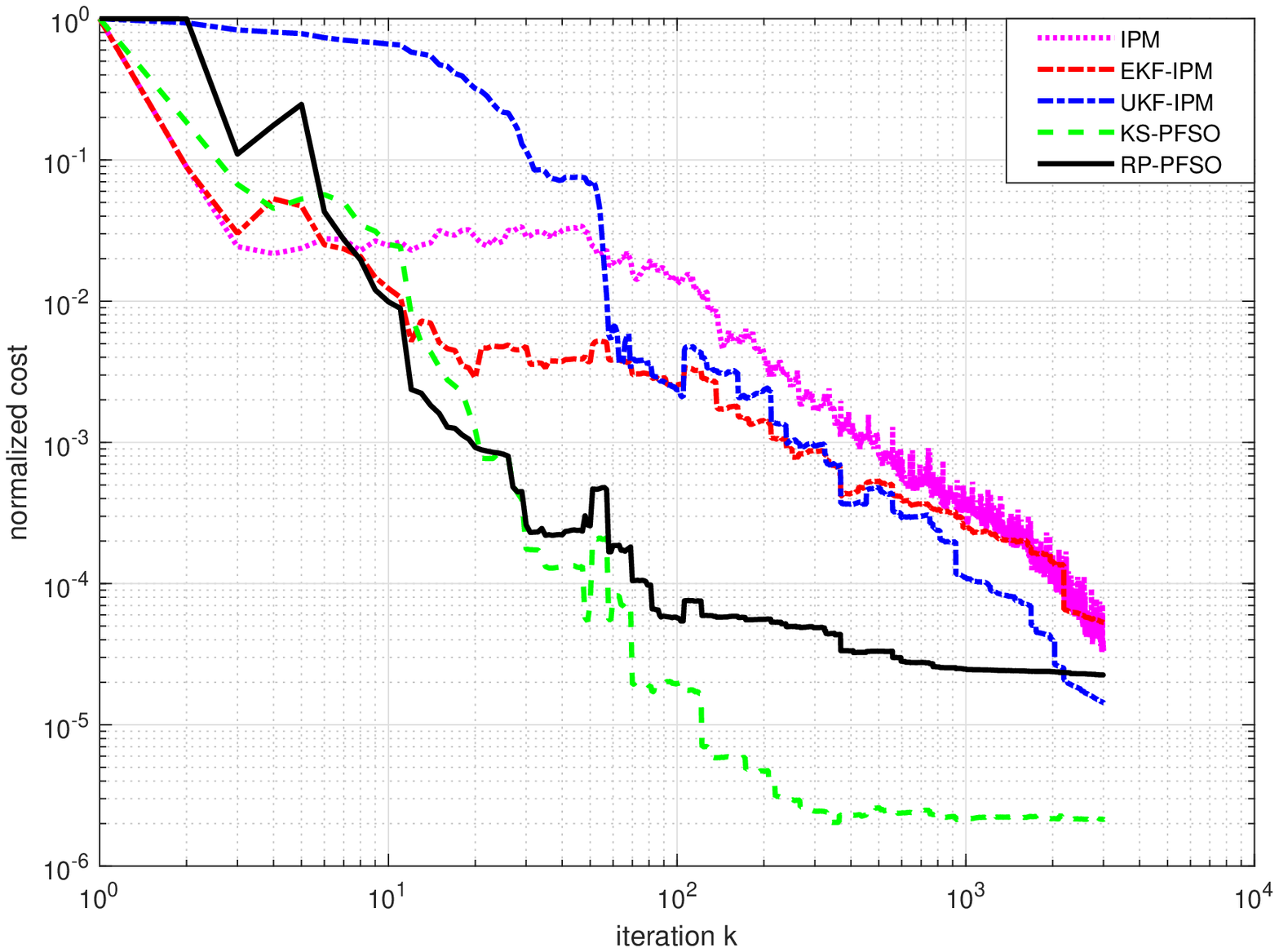}}\\
\centerline{\includegraphics[width=3.8in,height=2.5in]{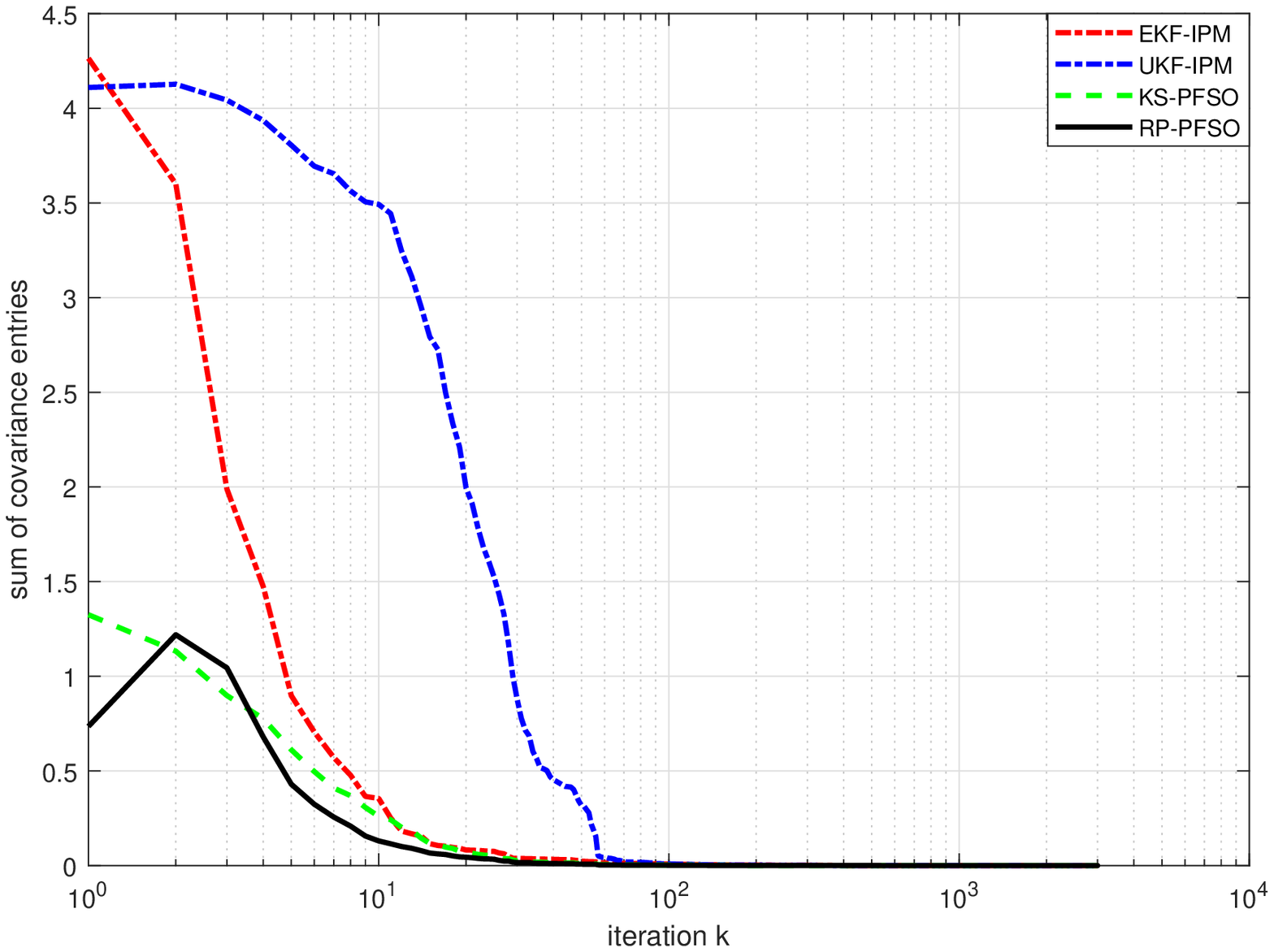}}
\end{tabular}
\caption{Fitting a more complex and highly nonlinear function as presented in subsection \ref{sec:park91a}.
Each algorithm is run 30 times independently. This figure shows the result averaged over these runs.
The upper panel shows the evolution of the normalized cost over the iterations.
The lower panel depicts the sum of the absolute values of the entries of $\hat{\mathbf{V}}$ per iteration.
One can see that KF-PFSO performs best, KS-PFSO and RP-PFSO converge faster than the others,
and the vanilla IPM suffers again from instability in the final phase of the iterations.} \label{fig:simu_case2}
\end{figure}

We conducted another least-square fitting experiment to further test our methods.
The experimental setting is the same as in subsection \ref{sec:sigmoid} except for
the formulation of $h_k(\theta)$, which is now defined to be:
\begin{equation}
h_k(\theta)=\frac{\theta_{(1)}}{x_{k,(1)}}\left[\sqrt{1+\left(\theta_{(2)}+\theta_{(3)}^{2}\right)
\frac{\theta_{(4)}}{\theta_{(1)}^{2}}}-x_{k,(2)}\right]+\left(\theta_{(1)}+x_{k,(3)} \theta_{(4)}\right)
\exp \left[x_{k,(4)}+\sin \left(\theta_{(3)}\right)\right],
\end{equation}
where $\theta\triangleq[\theta_{(1)},\theta_{(2)},\theta_{(3)}, \theta_{(4)}]^T$,
$x_k\triangleq[x_{k,(1)},x_{k,(2)},x_{k,(3)}, x_{k,(4)}]^T$. Each algorithm is run 30 times independently.
The averaged $NC(k)$, $k=0,\ldots,K$, over these runs is calculated and plotted in Fig.\ref{fig:simu_case2}. A tabular show of Fig.\ref{fig:simu_case2}
is presented in Tables \ref{Table:top_fig3}-Tables \ref{Table:bottom_fig3}. We see that,
for this highly nonlinear case, KS-PFSO performs significantly better than the other competitors. The vanilla IPM again suffers from instability during the final iterative phase.
\begin{table*}[!phtb]\caption{A tabular show for the top panel of Fig.\ref{fig:simu_case2}}\label{Table:top_fig3}
\centering
\begin{tabular}{cccccccc }
    \hline $k$ & 1 & 4    & 7    &  10    & 100             & 1000               & 3000\\
    \hline IPM & 1 &0.022 &0.028 & 0.025 & 0.013            & $3.0\times10^{-4}$ & $3.3\times10^{-5}$ \\
\hline EKF-IPM & 1 &0.053 &0.024 & 0.012 & 0.003            & $2.5\times10^{-4}$ & $5.3\times10^{-5}$\\
\hline UKF-IPM & 1 &0.804 &0.705 & 0.662 & 0.002            & $1.1\times10^{-4}$ & $1.4\times10^{-5}$\\
\hline KS-PFSO & 1 &0.454 &0.050 & 0.025 & $2\times10^{-5}$ & $2.2\times10^{-6}$ & $2.1\times10^{-6}$\\
\hline RP-PFSO & 1 &0.176 &0.028 & 0.010 & $6\times10^{-5}$ & $2.5\times10^{-5}$ & $2.3\times10^{-5}$\\
\hline
\end{tabular}
\end{table*}
\begin{table*}[!phtb]\caption{A tabular show for the bottom panel of Fig.\ref{fig:simu_case2}}\label{Table:bottom_fig3}
\centering
\begin{tabular}{cccccccc }
    \hline $k$ & 1     & 4    & 7    &  10    & 100   & 1000               & 3000\\
\hline EKF-IPM & 4.253 &1.475 &0.571 & 0.355  & 0.008 & $2.5\times10^{-4}$ & $4.1\times10^{-5}$\\
\hline UKF-IPM & 4.110 &3.936 &3.655 & 3.492  & 0.009 & $2.3\times10^{-4}$ & $3.6\times10^{-5}$\\
\hline KS-PFSO & 1.325 &0.777 &0.410 & 0.260  & 0.003 & $5.4\times10^{-5}$ & $7.7\times10^{-6}$\\
\hline RP-PFSO & 0.735 &0.680 &0.257 & 0.130  & 0.002 & $4.6\times10^{-5}$ & $4.6\times10^{-6}$\\
\hline
\end{tabular}
\end{table*}
\subsection{Binary classification using real data sets}
We then tested the proposed PFSO methods with 6 UCI data sets \cite{blake1998uci}: Haberman \cite{downs2001exact},
HTRU2 \cite{lyon2016fifty}, IRIS, Banknote Authentication, Pima Indians Diabetes, Skin Segmentation. Table \ref{Table:uci} summarizes the data sets. In our tests, the vanilla IPM, UKF-IPM,
and a typical SGD algorithm termed Adaline \cite{bottou1998online} are included for performance comparison.
We focused on ERM for binary classification. The objective function is (\ref{eqn:obj_fun}), where $y\in\{1,-1\}$ is the label and $\mathbf{x}$ the
feature vector. Except IRIS, each data set consists of two classes of data
instances. IRIS has three classes. We covert IRIS into a two-class data set by combining its first two classes into one.
We considered two types of loss functions, namely the least-quadratic (LQ) function $f_k(\theta)\triangleq(y_k-h_k(\theta))^2$,
where $h_k(\theta)=\frac{1}{1+\exp(-\alpha-\beta^{\top}\mathbf{x}_k)}$, and the logistic function
$f_k(\theta)\triangleq\log(1+\exp(-y(\alpha+\beta^{\top}\mathbf{x}_k)))$. For both of them, we have $\theta\triangleq[\alpha,\beta^{\top}]^{\top}$.
\begin{table*}[!phtb]\caption{Benchmark datsets from UCI}\label{Table:uci}
\centering
\begin{tabular}{ccccccc }
\hline Name of data set & Haberman & HTRU2 & IRIS & Banknote & Pima & Skin\\
\hline Number of Instances & 306 & 17898 & 150 & 1372 & 768  & 245057 \\
\hline Number of Attributes & 3 & 9 & 4 & 5 & 8 & 4\\
\hline
\end{tabular}
\end{table*}
\begin{table*}[!phtb]\caption{Error rate results of the algorithms when applied to analyze the UCI benchmark data sets. The best result given by these algorithms is indicated with boldface font.}\label{Table:res_uci}
\centering
%\begin{tabular}{c|c||c|c|c|c|c|c }\hline
\begin{tabular}{cccccccc }\hline
%\multicolumn{2}{c}{Algorithm} & Haberman & HTRU2 & IRIS & Banknote & Pima & Skin\\
Algorithm & Loss function & Haberman & HTRU2 & IRIS & Banknote & Pima & Skin\\ \hline
 Adaline & LQ & 0.2647 & 0.0899 & 0.3333 & 0.3943 & 0.3490 & 0.2187 \\ \hline
 IPM & LQ & 0.2647 & 0.0514 & 0.3333 & 0.0255 & 0.3490 & 0.0600 \\ \hline
 EKF-IPM & LQ & 0.2647 & 0.0514 & 0.3267 & 0.0445 & 0.3490 & 0.0849\\ \hline
 UKF-IPM & LQ & 0.2647 & 0.0514 & 0.3333 & 0.0452 & 0.3490 & 0.0845\\ \hline
 \multirow{2}*{KS-PFSO} & LQ &0.2647 & 0.0447 & 0.0933 & \textbf{0.0233} & 0.3060 & \textbf{0.0592} \\\cmidrule(l){2-8}
  & Logistic &\textbf{0.2549} & \textbf{0.0363} & \textbf{0.0533} & 0.0561 & \textbf{0.2708} & 0.1110 \\ \hline
 \multirow{2}*{RP-PFSO}  & LQ & 0.2647 & \textbf{0.0363} & 0.1000 & 0.0241 & 0.3021 & 0.0726 \\\cmidrule(l){2-8}
  & Logistic & 0.2582 & 0.0397 & 0.0533 & 0.0437 & 0.2839 & 0.1098 \\ \hline
\end{tabular}
\end{table*}
\begin{table*}[!phtb]\caption{Computing time comparison (the unit of the time is 3 milliseconds). This experiment is conducted with a one-core Intel i5-3210M 2.50 GHz processor. The particle size for all PFSOs is set at 1000. }\label{Table:time_uci}
\centering
\begin{tabular}{ccccccc}\hline
Algorithm &  Adaline & IPM & EKF-IPM & UKF-IPM & KS-PFSO & RP-PFSO\\ \hline
Scaled time & 1 & 2.6686 & 2.6544 & 21.7107& $2.1500\times10^3$ & $3.7313\times10^3$ \\ \hline
\end{tabular}
\end{table*}
For performance evaluation, we adopt the 10-fold cross-validation technique. Given the optimum $\theta^{\star}\triangleq[\alpha^{\star},(\beta^{\star})^{\top}]^{\top}$ that has been found,
we predict the label of a data instance $\{\mathbf{x},y\}$ by checking the value of $\frac{1}{1+\exp(-\alpha^{\star}-(\beta^{\star})^{\top}\mathbf{x})}$.
If it is bigger than 0.5, then we set the predicted label to 1; otherwise to -1.
By comparing the predicted labels with the ground truth for data items in the test set, one can obtain the error rate, i.e., the performance metric adopted here.

For each data set, we do the same initialization for each algorithm. The regularization parameter $\lambda$ is set
at 0.25. The particle size of PFSOs is $N=4\times 10^3$. All the other parameters are initialized in the same way
as that presented in Subsection \ref{sec:toy_data}. The learning rate parameter in Adaline is set to $1/k$,
where $k$ denotes the iteration index.

\begin{figure}[!htb]
\begin{tabular}{c}
\centerline{\includegraphics[width=3.8in,height=2.8in]{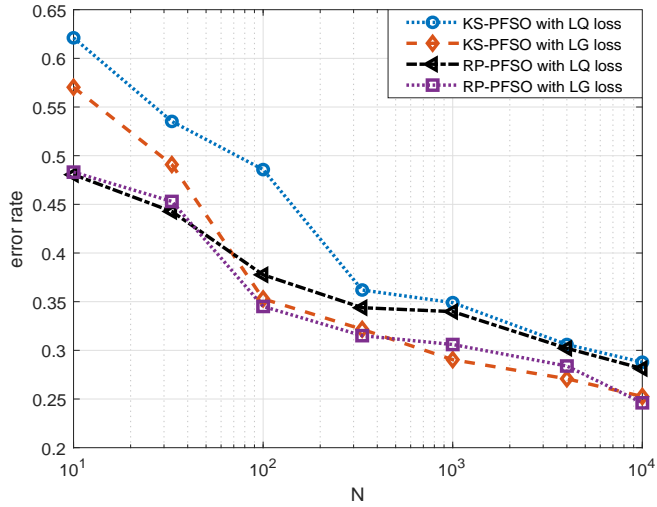}}
\end{tabular}
\caption{The particle size $N$ of the PFSOs vs. the error rate.
This experiment is conducted using the UCI benchmark data set termed Pima-Indians-diabetes.
The abbreviations ``LQ"  and ``LG" represent ``least-quadratic" and ``Logistic", respectively.} \label{fig:classification_N_vs_error}
\end{figure}

The experimental result is presented in Table \ref{Table:res_uci}. It shows that, for every data set, the best classification result in terms of error rate is always given by KS-PFSO. For datasets Haberman, HTRU2, IRIS and Pima, the employment of the logistic loss function leads to better performance than of the LQ function, in terms of error rate.
For the other two data sets, Banknote and Skin, the LQ type loss function is preferable to
the logistic loss function for use with KS-PFSO. Besides, we see that the performance of UKF-IPM
is indistinguishable from that of EKF-IPM. We also checked the influence
of the particle size $N$ on the error rate for PFSOs, which is shown in Fig.\ref{fig:classification_N_vs_error}.
Once again we see that RP-PFSO outperforms KS-PFSO when $N$ takes small values.
\subsection{Summary of Results}
Overall, for all cases considered here, the proposed PFSO method shows a promising level of performance in terms of convergence rate, as anticipated. Its flexibility in handling different types of loss functions is also verified. That says PFSO finds more applications than KF-IPMs. The particle size $N$ is a major factor that influences the error rate of PFSO. When $N$ takes small values, RP-PFSO performs better than KS-PFSO, which shows the evidence of the benefit provided by the RP operation.

We also compared the involved algorithms in terms of computing time. As shown in Table \ref{Table:time_uci}, the PFSO methods consume more computing time, while it is worth noting that PFSO can be markedly accelerated by algorithmic parallelization. Specifically, the particle generation and the importance weighting steps can be parallelized straightforward, as they only involve independent operations on each particle. The resampling procedure can also be parallelized as reported in \cite{murray2016parallel}. The Rao-Blackwellization technique can provide another way to accelerate the PF methods if there is any analytic structure included in the model \cite{liu1998sequential,sarkka2007rao}.
\section{Concluding Remarks}\label{sec:conclusions}
Motivated by both the success and the limitations of the KF-type IPM (Section \ref{sec:background}),
we proposed a PF based IPM scheme called PFSO for large-scale stochastic optimization. We presented two specific implementations of PFSO, termed KS-PFSO and RP-PFSO. The PF method in itself is not new since it has been developed for more
than two decades since the seminal paper \cite{gordon1993novel} was published, while, the PFSO methods presented here are novel as they are distinctive from any existent stochastic optimization method.
Specifically, our PFSO methods own four desirable properties: 
\begin{itemize}
\item they free the user from computing
gradients (actually they do not require the objective function be differentiable); 
\item they have no requirement
on the form of the components of the objective function; 
\item they outperforms their competitors significantly in terms of convergence rate and accuracy; 
\item their performance can be easily tuned by adapting the particle size.
\end{itemize}
The performance of the PFSO methods has been empirically demonstrated based on both simulated and real-life data sets.

This work opens the door to borrow a rich body of PF methods to solve large-scale stochastic optimization problems.
For example, one can borrow ideas from adaptive importance sampling, see e.g., in \cite{cappe2008adaptive,oh1992adaptive,liu2014adaptive},
to design adaptive PFSO methods that can build up the proposal functions automatically. Our recent work on robust
PF \cite{liu2017ilapf,liu2017robust,dai2016robust,liu2011instantaneous,liu2019robust} may also provide tools for developing robust PFSO methods.

By far a rigorous theoretical analysis for PFSO lacks, while the theoretical work on PF and Sequential Monte Carlo is rich, see e.g., in \cite{crisan2002survey,sarkka2013bayesian,douc2014long,hu2008basic,hu2011general}, and can provide clues for analyzing PFSO.
Besides, another future research direction is to verify the robustness of our methods by evaluating them on data sets of larger scales.
\bibliographystyle{IEEEbib}
\bibliography{mybibliography}

\begin{thebibliography}{10}

\bibitem{reddi2017new}
S.J. Reddi,
\newblock {\em New Optimization Methods for Modern Machine Learning},
\newblock Ph.D. thesis, Carnegie Mellon University, 2017.

\bibitem{bottou2018optimization}
L.~Bottou, F.E. Curtis, and J.~Nocedal,
\newblock ``Optimization methods for large-scale machine learning,''
\newblock {\em SIAM Review}, vol. 60, no. 2, pp. 223--311, 2018.

\bibitem{bottou2010large}
L.~Bottou,
\newblock ``Large-scale machine learning with stochastic gradient descent,''
\newblock in {\em Proceedings of COMPSTAT'2010}, pp. 177--186. Springer, 2010.

\bibitem{bottou1998online}
L.~Bottou,
\newblock ``Online learning and stochastic approximations,''
\newblock {\em On-line learning in neural networks}, vol. 17, no. 9--42, pp.
  142, 1998.

\bibitem{qian1999momentum}
N.~Qian,
\newblock ``On the momentum term in gradient descent learning algorithms,''
\newblock {\em Neural Networks}, vol. 12, no. 1, pp. 145--151, 1999.

\bibitem{nesterov1983method}
Y.~Nesterov,
\newblock ``A method for unconstrained convex minimization problem with the
  rate of convergence o (1/k\^{} 2),''
\newblock in {\em Doklady ANSSSR}, 1983, vol. 269, pp. 543--547.

\bibitem{duchi2011adaptive}
J.~Duchi, E.~Hazan, and Y.~Singer,
\newblock ``Adaptive subgradient methods for online learning and stochastic
  optimization,''
\newblock {\em Journal of Machine Learning Research}, vol. 12, no. Jul, pp.
  2121--2159, 2011.

\bibitem{zeiler2012adadelta}
M.D. Zeiler,
\newblock ``Adadelta: an adaptive learning rate method,''
\newblock {\em arXiv preprint arXiv:1212.5701}, 2012.

\bibitem{kingma2014adam}
D.~P. Kingma and J.~Ba,
\newblock ``Adam: A method for stochastic optimization,''
\newblock in {\em Int'l Conf. on Learning Representations (ICLR)}, 2015, pp.
  1--13.

\bibitem{parikh2014proximal}
N.~Parikh and S.~Boyd,
\newblock ``Proximal algorithms,''
\newblock {\em Foundations and Trends{\textregistered} in Optimization}, vol.
  1, no. 3, pp. 127--239, 2014.

\bibitem{bertsekas2011incremental}
D.P. Bertsekas,
\newblock ``Incremental proximal methods for large scale convex optimization,''
\newblock {\em Mathematical programming}, vol. 129, no. 2, pp. 163, 2011.

\bibitem{aky2018the}
{\"O}.D. Aky{\i}ld{\i}z, V.~Elvira, and J.~M{\'\i}guez,
\newblock ``The incremental proximal method: A probabilistic perspective,''
\newblock in {\em Proc. of IEEE Int'l Conf. on Acoustics, Speech, and Signal
  Processing (ICASSP)}. IEEE, 2018, pp. 4279--4283.

\bibitem{gharaei2019joint}
A.~Gharaei, M.~Karimi, and Seyed~A. Hoseini~S.,
\newblock ``Joint economic lot-sizing in multi-product multi-level integrated
  supply chains: generalized benders decomposition,''
\newblock {\em International Journal of Systems Science: Operations \&
  Logistics}, pp. 1--17, 2019.

\bibitem{gharaei2019integrated}
A.~Gharaei, M.~Karimi, and S.~Shekarabi,
\newblock ``An integrated multi-product, multi-buyer supply chain under
  penalty, green, and quality control polices and a vendor managed inventory
  with consignment stock agreement: The outer approximation with equality
  relaxation and augmented penalty algorithm,''
\newblock {\em Applied Mathematical Modelling}, vol. 69, pp. 223--254, 2019.

\bibitem{duan2018selective}
C.~Duan, C.~Deng, A.~Gharaei, J.~Wu, and B.~Wang,
\newblock ``Selective maintenance scheduling under stochastic maintenance
  quality with multiple maintenance actions,''
\newblock {\em International Journal of Production Research}, vol. 56, no. 23,
  pp. 7160--7178, 2018.

\bibitem{hoseini2019modelling}
Seyed~A. Hoseini~S., A.~Gharaei, and M.~Karimi,
\newblock ``Modelling and optimal lot-sizing of integrated multi-level
  multi-wholesaler supply chains under the shortage and limited warehouse
  space: generalised outer approximation,''
\newblock {\em International Journal of Systems Science: Operations \&
  Logistics}, vol. 6, no. 3, pp. 237--257, 2019.

\bibitem{gharaei2019modelling}
A.~Gharaei, Seyed~A. Hoseini~S., and M.~Karimi,
\newblock ``Modelling and optimal lot-sizing of the replenishments in
  constrained, multi-product and bi-objective epq models with defective
  products: Generalised cross decomposition,''
\newblock {\em International Journal of Systems Science: Operations \&
  Logistics}, pp. 1--13, 2019.

\bibitem{gharaei2019integrated2}
A.~Gharaei, Seyed~A. Hoseini~S., M.~Karimi, E.~Pourjavad, and A.~Amjadian,
\newblock ``An integrated stochastic epq model under quality and green
  policies: generalised cross decomposition under the separability approach,''
\newblock {\em International Journal of Systems Science: Operations \&
  Logistics}, pp. 1--13, 2019.

\bibitem{haykin2001kalman}
S.~Haykin,
\newblock {\em Kalman filtering and neural networks},
\newblock Wiley Online Library, 2001.

\bibitem{arulampalam2002tutorial}
M.~S. Arulampalam, S.~Maskell, N.~Gordon, and T.~Clapp,
\newblock ``A tutorial on particle filters for online nonlinear/non-{G}aussian
  {B}ayesian tracking,''
\newblock {\em IEEE Trans. on Signal Processing}, vol. 50, no. 2, pp. 174--188,
  2002.

\bibitem{van2001unscented}
R.~Van Der~Merwe, A.~Doucet, N.~De~Freitas, and E.A. Wan,
\newblock ``The unscented particle filter,''
\newblock in {\em Advances in Neural Information Processing Systems (NIPS)},
  2001, pp. 584--590.

\bibitem{liu2008particle}
B.~Liu, X.~Ma, and C.~Hou,
\newblock ``A particle filter using {SVD} based sampling {K}alman filter to
  obtain the proposal distribution,''
\newblock in {\em Proc. of IEEE Conf. on Cybernetics and Intelligent Systems}.
  IEEE, 2008, pp. 581--584.

\bibitem{liu2001combined}
J.~Liu and M.~West,
\newblock ``Combined parameter and state estimation in simulation-based
  filtering,''
\newblock in {\em Sequential Monte Carlo methods in practice}, pp. 197--223.
  Springer, 2001.

\bibitem{gordon1993novel}
N.~Gordon, D.~Salmond, and A.~F.~M. Smith,
\newblock ``Novel approach to nonlinear/non-{G}aussian {B}ayesian state
  estimation,''
\newblock {\em IEE Proceedings F (Radar and Signal Processing)}, vol. 140, no.
  2, pp. 107--113, 1993.

\bibitem{liu2008single}
B.~Liu, C.~Ji, X.~Ma, and C.~Hou,
\newblock ``Single-tone frequency tracking using a particle filter with
  improvement strategies,''
\newblock in {\em Int'l Conf. on Audio, Language and Image Processing
  (ICALIP)}. IEEE, 2008, pp. 1615--1619.

\bibitem{gilks2001following}
W.~R. Gilks and C.~Berzuini,
\newblock ``Following a moving target--{M}onte {C}arlo inference for dynamic
  {B}ayesian models,''
\newblock {\em Journal of the Royal Statistical Society: Series B (Statistical
  Methodology)}, vol. 63, no. 1, pp. 127--146, 2001.

\bibitem{chopin2002sequential}
N.~Chopin,
\newblock ``A sequential particle filter method for static models,''
\newblock {\em Biometrika}, vol. 89, no. 3, pp. 539--552, 2002.

\bibitem{ridgeway2003sequential}
G.~Ridgeway and D.~Madigan,
\newblock ``A sequential {M}onte {C}arlo method for {B}ayesian analysis of
  massive datasets,''
\newblock {\em Data Mining and Knowledge Discovery}, vol. 7, no. 3, pp.
  301--319, 2003.

\bibitem{patel2016kalman}
V.~Patel,
\newblock ``Kalman-based stochastic gradient method with stop condition and
  insensitivity to conditioning,''
\newblock {\em SIAM Journal on Optimization}, vol. 26, no. 4, pp. 2620--2648,
  2016.

\bibitem{bell1993iterated}
B.M. Bell and F.W. Cathey,
\newblock ``The iterated {K}alman filter update as a {G}auss-{N}ewton method,''
\newblock {\em IEEE Trans. on Automatic Control}, vol. 38, no. 2, pp. 294--297,
  1993.

\bibitem{bertsekas1996incremental}
D.P. Bertsekas,
\newblock ``Incremental least squares methods and the extended kalman filter,''
\newblock {\em SIAM Journal on Optimization}, vol. 6, no. 3, pp. 807--822,
  1996.

\bibitem{ho1962stochastic}
Y.C. Ho,
\newblock ``On the stochastic approximation method and optimal filtering
  theory,''
\newblock {\em Journal of Mathematical Analysis and Applications}, vol. 6, pp.
  152--154, 1962.

\bibitem{stinis2012stochastic}
P.~Stinis,
\newblock ``Stochastic global optimization as a filtering problem,''
\newblock {\em Journal of Computational Physics}, vol. 231, no. 4, pp.
  2002--2014, 2012.

\bibitem{liu2017posterior}
B.~Liu,
\newblock ``Posterior exploration based sequential {M}onte {C}arlo for global
  optimization,''
\newblock {\em Journal of Global Optimization}, vol. 69, no. 4, pp. 847--868,
  2017.

\bibitem{liu2016particle}
B.~Liu, S.~Cheng, and Y.~Shi,
\newblock ``Particle filter optimization: A brief introduction,''
\newblock in {\em Int'l Conf. in Swarm Intelligence}. Springer, 2016, pp.
  95--104.

\bibitem{del2006sequential}
P.~Del~Moral, A.~Doucet, and A.~Jasra,
\newblock ``Sequential {M}onte {C}arlo samplers,''
\newblock {\em Journal of the Royal Statistical Society: Series B (Statistical
  Methodology)}, vol. 68, no. 3, pp. 411--436, 2006.

\bibitem{julier2004unscented}
S.J. Julier and J.K. Uhlmann,
\newblock ``Unscented filtering and nonlinear estimation,''
\newblock {\em Proceedings of the IEEE}, vol. 92, no. 3, pp. 401--422, 2004.

\bibitem{blake1998uci}
C.L. Blake,
\newblock ``Uci repository of machine learning databases,''
\newblock {\em http://www.ics.uci.edu/\~{}mlearn/MLRepository.html}, 1998.

\bibitem{downs2001exact}
T.~Downs, K.E. Gates, and A.~Masters,
\newblock ``Exact simplification of support vector solutions,''
\newblock {\em Journal of Machine Learning Research}, vol. 2, pp. 293--297,
  2001.

\bibitem{lyon2016fifty}
R.~J. Lyon, B.~W. Stappers, S.~Cooper, J.~M. Brooke, and J.~D. Knowles,
\newblock ``Fifty years of pulsar candidate selection: from simple filters to a
  new principled real-time classification approach,''
\newblock {\em Monthly Notices of the Royal Astronomical Society}, vol. 459,
  no. 1, pp. 1104--1123, 2016.

\bibitem{murray2016parallel}
L.M. Murray, A.~Lee, and P.E. Jacob,
\newblock ``Parallel resampling in the particle filter,''
\newblock {\em Journal of Computational and Graphical Statistics}, vol. 25, no.
  3, pp. 789--805, 2016.

\bibitem{liu1998sequential}
J.S. Liu and R.~Chen,
\newblock ``Sequential {M}onte {C}arlo methods for dynamic systems,''
\newblock {\em Journal of the American statistical association}, vol. 93, no.
  443, pp. 1032--1044, 1998.

\bibitem{sarkka2007rao}
S.~S{\"a}rkk{\"a}, A.~Vehtari, and J.~Lampinen,
\newblock ``Rao-blackwellized particle filter for multiple target tracking,''
\newblock {\em Information Fusion}, vol. 8, no. 1, pp. 2--15, 2007.

\bibitem{cappe2008adaptive}
O.~Capp{\'e}, R.~Douc, A.~Guillin, J.~Marin, and C.P. Robert,
\newblock ``Adaptive importance sampling in general mixture classes,''
\newblock {\em Statistics and Computing}, vol. 18, no. 4, pp. 447--459, 2008.

\bibitem{oh1992adaptive}
M.~Oh and J.O. Berger,
\newblock ``Adaptive importance sampling in {M}onte {C}arlo integration,''
\newblock {\em Journal of Statistical Computation and Simulation}, vol. 41, no.
  3-4, pp. 143--168, 1992.

\bibitem{liu2014adaptive}
B.~Liu,
\newblock ``Adaptive annealed importance sampling for multimodal posterior
  exploration and model selection with application to extrasolar planet
  detection,''
\newblock {\em The Astrophysical Journal Supplement Series}, vol. 213, no. 14,
  pp. 1--16, 2014.

\bibitem{liu2017ilapf}
B.~Liu,
\newblock ``{ILAPF}: Incremental learning assisted particle filtering,''
\newblock in {\em Proc. of IEEE Int'l Conf. on Acoustics, Speech and Signal
  Processing (ICASSP 2018)}, 2018, pp. 4284--4288.

\bibitem{liu2017robust}
B.~Liu,
\newblock ``Robust particle filter by dynamic averaging of multiple noise
  models,''
\newblock in {\em Proc. of the 42nd IEEE Int'l Conf. on Acoustics, Speech, and
  Signal Processing (ICASSP)}. IEEE, 2017, pp. 4034--4038.

\bibitem{dai2016robust}
Y.~Dai and B.~Liu,
\newblock ``Robust video object tracking via {B}ayesian model averaging-based
  feature fusion,''
\newblock {\em Optical Engineering}, vol. 55, no. 8, pp. 1--11, 2016.

\bibitem{liu2011instantaneous}
B.~Liu,
\newblock ``Instantaneous frequency tracking under model uncertainty via
  dynamic model averaging and particle filtering,''
\newblock {\em IEEE Trans. on Wireless Communications}, vol. 10, no. 6, pp.
  1810--1819, 2011.

\bibitem{liu2019robust}
B.~Liu,
\newblock ``Robust particle filtering via {B}ayesian nonparametric outlier
  modeling,''
\newblock in {\em Int'l Conf. on Information Fusion (FUSION)}, 2019, pp.
  102--106.

\bibitem{crisan2002survey}
D.~Crisan and A.~Doucet,
\newblock ``A survey of convergence results on particle filtering methods for
  practitioners,''
\newblock {\em IEEE Trans. on signal processing}, vol. 50, no. 3, pp. 736--746,
  2002.

\bibitem{sarkka2013bayesian}
S.~S{\"a}rkk{\"a},
\newblock {\em Bayesian filtering and smoothing}, vol.~3,
\newblock Cambridge University Press, 2013.

\bibitem{douc2014long}
R.~Douc, E.~Moulines, and J.~Olsson,
\newblock ``Long-term stability of sequential monte carlo methods under
  verifiable conditions,''
\newblock {\em The Annals of Applied Probability}, vol. 24, no. 5, pp.
  1767--1802, 2014.

\bibitem{hu2008basic}
X.~Hu, T.~Sch{\"o}n, and L.~Ljung,
\newblock ``A basic convergence result for particle filtering,''
\newblock {\em IEEE Transactions on Signal Processing}, vol. 56, no. 4, pp.
  1337--1348, 2008.

\bibitem{hu2011general}
X.~Hu, T.~Sch{\"o}n, and L.~Ljung,
\newblock ``A general convergence result for particle filtering,''
\newblock {\em IEEE Trans. on Signal Processing}, vol. 59, no. 7, pp.
  3424--3429, 2011.

\end{thebibliography}
\end{document}